\documentclass{article}

\usepackage[english]{babel}

\usepackage[letterpaper,top=2cm,bottom=2cm,left=3cm,right=3cm,marginparwidth=1.75cm]{geometry}

\usepackage{parskip}
\usepackage{graphicx}
\usepackage[colorlinks=true, allcolors=blue]{hyperref}
\usepackage{quoting}
\usepackage{mdframed}
\usepackage{graphicx}
\usepackage{booktabs}
\usepackage{amsmath}
\usepackage{array}
\usepackage{enumitem}
\usepackage[natbibapa]{apacite}
\usepackage{appendix}
\usepackage{hyperref}

\newcolumntype{M}[1]{>{\centering\arraybackslash}p{#1}}

\title{EQ-Bench: An Emotional Intelligence Benchmark for Large Language Models}
\author{Samuel J. Paech}
\date{December 11, 2023}

\begin{document}
\maketitle

\begin{abstract}
We introduce EQ-Bench, a novel benchmark designed to evaluate aspects of emotional intelligence in Large Language Models (LLMs). We assess the ability of LLMs to understand complex emotions and social interactions by asking them to predict the intensity of emotional states of characters in a dialogue. The benchmark is able to discriminate effectively between a wide range of models. We find that EQ-Bench correlates strongly with comprehensive multi-domain benchmarks like MMLU \citep{hendrycks2020measuring} (r=0.97), indicating that we may be capturing similar aspects of broad intelligence. Our benchmark produces highly repeatable results using a set of 60 English-language questions. We also provide open-source code for an automated benchmarking pipeline at \href{https://github.com/EQ-bench/EQ-Bench}{https://github.com/EQ-bench/EQ-Bench} and a leaderboard at \href{https://eqbench.com}{https://eqbench.com}.
\end{abstract}

\section{Introduction}

Emotional intelligence (EI or, informally, EQ) is a cornerstone of human cognition, influencing every-thing from decision-making to interpersonal interactions \citep{goleman1996emotional}. Pioneers in the field of emotional intelligence, \citet{salovey1990emotional} define EI as "The ability to monitor one's own and others' feelings, to discriminate among them, and to use this information to guide one's thinking and action." This was later broken down into four branches: perceiving emotions (non-verbally), using emotions, understanding emotions and managing emotions \citep{mayer1997emotional}.

Some multi-modal LLMs such as GPT-4 \citep{openai2023gpt4} have demonstrated capabilities in several branches of EI \citep{yang2023dawn}. However, our work will focus on emotional understanding (EU): The ability to comprehend and interpret complex emotions and their meanings in social contexts. This branch of EI is most suited for assessing LLMs that only operate in the text modality. Emotional and social understanding is important for language models, since they primarily interact with humans via a natural language conversation. The ability to comprehend the language of emotion, and to grasp the complexities and nuances of emotional interactions, is a fundamental part of cognition and intelligence more broadly.

Existing benchmarks assess LLM capabilities in different ways: Some assess a range of knowledge domains; others focus on specific areas, like coding ability; others compare models based on preferred text output. The existing industry standard benchmarks for LLMs do not specifically target EU, and there is likewise a lack of such tests described in the literature, with the notable exception of SECEU \citep{wang2023emotional}. This may be because EU is not trivial to assess in language models with an objective metric. Previous work has shown ChatGPT to outperform humans in emotional awareness \citep{elyoseph2023chatgpt}, however the scoring of such tests typically requires interpretation by experts (in this case, licensed psychologists).  We aim to fill this gap with EQ-Bench, by improving on the question format introduced by SECEU to create a benchmark that effectively measures EU in LLMs.

This paper presents a novel benchmark that assesses a LLM's understanding of emotions and interpersonal dynamics. To assess EU, we focus on a specific question format: rating the emotional intensity of characters in a GPT-4 generated dialogue depicting a scene of conflict or tension. This question format allows for a more challenging and nuanced assessment of the subject's understanding than is typically possible with a multiple choice question, while avoiding the need for interpretation by an assessor. The focus on scenes of conflict and tension (in either a positive or negative context) is chosen for its efficacy at producing nuanced emotional inflection within a short dialogue. We find that despite the relatively narrow focus of the dialogue scenarios, EQ-Bench differentiates model capabilities effectively, and results correlate strongly with industry standard LLM benchmarks that assess broad intelligence.

\begin{figure}[htbp]
\centering
\includegraphics[width=280px]{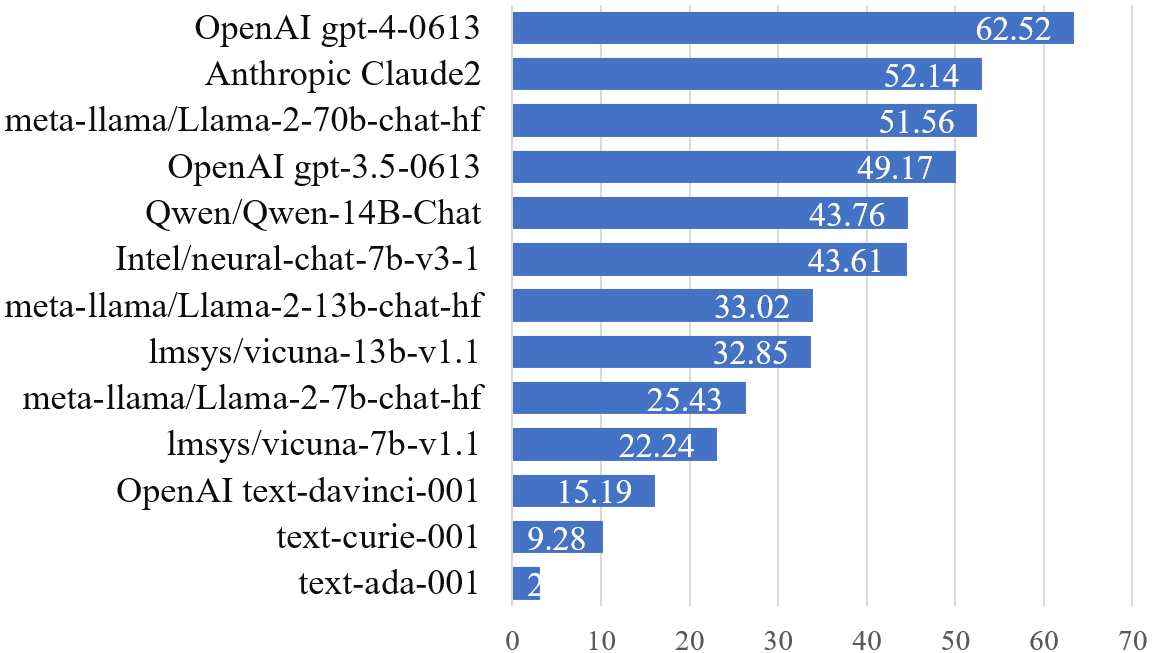}
\caption{Chart of EQ-Bench scores for a subset of models tested (for full list see \autoref{tab:eqbench}).}
\label{fig:bars}
\end{figure}

\section{Background and Motivation}

\subsection{Existing Psychometrics Tests for Emotional Intelligence}

Many tests for EI have been developed for humans in such a way that make them unsuitable for assessing LLMs. Aspects of these tests which are poorly suited to a LLM benchmark include: self-rating of abilities; multi-modal questions (i.e. pictures); incorporating feedback from colleagues; questions specifically designed to test for cognitive disorders. Many of these tests are also closed source and require a fee or an accredited assessor, which represents a barrier to those in the open-source LLM community who might wish to test the EI of their model.

\subsection{Desired Features of LLM Benchmarks}

The assessment of LLMs is distinct from human psychometrics, and poses unique challenges. Industry standard benchmarks typically assess holistic or broad intelligence. Desirable properties of a benchmark designed to test LLM broad intelligence include:
\begin{itemize}
\item Results correlate with perceived capabilities of the model in real world use.
\item Results correlate with industry standard multi-domain benchmarks.
\item Difficult to "game" or cheat the benchmark.
\item Answers scored objectively without need for interpretation.
\item Able to differentiate a wide range of intelligence levels, with a high ceiling.
\item Computable in a reasonable time.
\end{itemize}

\subsection{Existing Benchmarks for LLMs}
Of the popular benchmarks used to evaluate LLM performance (not explicitly testing EI), there are three main categories:

\begin{enumerate}
\item Multi-domain multiple-choice question-answering.
\item LLM-as-a-judge.
\item Ranking by human preference of output.
\end{enumerate}

Each of these approaches has its strengths and drawbacks. The multi-domain QA tests require administration of a large number of questions in order to assess broad intelligence; 14,079 questions in the case of MMLU \citep{hendrycks2020measuring}. This can be computationally prohibitive. Less comprehensive tests are more easily gamed by targeting the training set of a model to perform well on the specific domains the benchmark covers.

LLM-as-a-judge tests are inherently limited to the capabilities of the model tasked with judging answers, and their scores will reflect the biases of that model. For instance, the AlpacaEval leaderboards contain a warning that "GPT-4 may favor models with longer outputs and/or those that were fine-tuned on GPT-4 outputs." For the above mentioned reasons, these kinds of benchmarks are not always a reliable indicator of the general capabilities or perceived intelligence of a model.

\subsection{Emotional Intelligence as a Proxy for Broad Intelligence in LLMs}

The relationship between EI and IQ has been widely studied in humans, but remains contentious, with studies disagreeing about whether there is a positive, negative or no relationship \citep{ogurlu2021meta}. We hypothesise that the story is different for LLMs: that EI is closely correlated to the broad intelligence of the model. In humans, a complex interplay of genetic and environmental factors influence EI and IQ \citep{vernon2008behavioral}, \citep{mcrae2017genetic}, \citep{turkheimer2003socioeconomic}. In contrast, state-of-the-art LLMs almost exclusively share similar Transformer-based architectures \citep{vaswani2017attention}, and are trained on massive sets of data covering a wide range of material. Given these differences between humans and LLMs, we suggest that it is reasonable to expect that LLM performance on EI tests may scale closely with benchmarks that assess broad intelligence. Our results indicate that this is indeed the case, producing very strong Pearson correlations with MMLU (r=0.97), HellaSwag (r=0.91) and others (\autoref{fig:scatter}).

\subsection{SECEU: A Promising Benchmark for Emotional Understanding}

The SECEU benchmark, developed by \citet{wang2023emotional}, measures emotional understanding (EU) in language models using an elegant approach: A short scenario is presented, after which the subject is asked to rate the relative strength of four candidate emotions that the person in the scenario might be feeling. In contrast to a traditional multiple-choice question, this approach affords more scope for assessing nuanced comprehension of the emotions at play. Crucially, this question format can be scored objectively without requiring interpretation by an assessor.

However, the SECEU benchmark is somewhat limited in its ability to reliably differentiate the capabilities of models. For example, SECEU scored OpenAI's Babbage and Curie models at or above the human average, which appears inconsistent with the limited capabilities of these models. It placed Curie, an early generation 13 billion parameter model, very close in score (within 2.6\%) to GPT-3.5-turbo \citep{wang2023emotional}, which is a significantly larger and generally considered a more advanced model.

We believe these discrepancies are attributable to a number of factors:

1. \textit{Standardising to Human Average Scores:} The SECEU's reference answers were determined by taking the average of a human cohort's responses to the test questions. The stated intention was to leverage the collective intelligence of the crowd, however this approach has been shown to introduce bias \citep{huyghe2022scoring}. We suggest that this approach may have compressed the upper range of EU that the test can effectively measure to somewhere near to the human mean. This is evidenced by some of the least capable models like OpenAI's Curie scoring above the human average.

2. \textit{Question Complexity:} The SECEU test's questions may not be complex enough to effectively assess a wide range of emotional understanding.

3. \textit{Requiring Answers to Sum to 10:} A third issue we identify is that the questions in the SECEU test require the emotion intensity ratings to sum to 10. This is problematic for two reasons: Firstly, language models, especially smaller ones, often struggle to perform basic mathematics \citep{hendrycks2021measuring}. This limitation can lead to a large number of invalid results. Secondly, LLMs are constrained to reason sequentially as they produce text \citep{yao2023tree}. Effectively this means they will be locked into the numerical ratings of the earlier questions before the later ratings are considered and decided upon. This can skew responses as the later answers are constrained by how many points are left out of 10 to allocate.

4. \textit{Most plausible emotions:} SECEU test questions present "four of the most plausible emotions" the character would feel in response to the scenario \citet{wang2023emotional}. Choosing only plausible, similar emotions may introduce ambiguities about the correct relative intensity for each, reducing the question's discriminatory power for assessing EU.

\subsection{Changes to the SECEU Question Format}

We propose the following improvements to the SECEU question format, which are implemented in EQ-Bench:

\begin{enumerate}
\item \textbf{Reference answers not decided by the crowd.} Answers were carefully chosen by the authors, rather than synthesised from the average responses of a human cohort. This is to avoid the potential problem of limiting the upper range of EU that the test is able to measure.
\item \textbf{More complex scenarios.} We chose to base questions around an emotionally charged short dialogue rather than the dialogue-less descriptive scenarios used in SECEU questions. We believe that the interpretation of the emotions at play in an observed dialogue lends itself to a more nuanced analysis of emotional interplay, and the dialogues and questions in EQ-Bench were specifically chosen to reward careful reading and insightful thinking.
\item \textbf{Removing the summation requirement.} The SECEU benchmark questions require all four emotional intensity ratings to sum to 10. Instead, we prompt models to rate each emotion's intensity on a scale of 0-10. This circumvents the problem of sequential reasoning, whereby the first answers effectively dictate the allowable range of ratings for later answers. This change allows the model to focus on the emotional content without being hindered by the typical language model constraints.
\item \textbf{Emotions selection:} Instead of selecting the four most plausible emotions to be rated, we selected four diverse emotions with the specific intention of minimising ambiguities about their expected relative intensity given a careful reading of the text.
\end{enumerate}

\section{Methodology}

\subsection{Question Format}

All questions take the form of this example:

\begin{quote}

\textit{At the end of this dialogue, Jane would feel:}
\begin{itemize}[label={}]
    \item \textit{Surprised:}
    \item \textit{Confused:}
    \item \textit{Angry:}
    \item \textit{Forgiving:}
\end{itemize}
\end{quote}

Each of the four emotions are to be rated in intensity from 0 to 10. The selected emotions typically include some that are clearly wrong, some that are obvious, and some that require a nuanced understanding of the scene in order to rate them accurately.

\subsection{Dialogue Generation}

We leveraged OpenAI's GPT-4 \citep{openai2023gpt4} to generate the dialogues to serve as context for the test questions. The reference answers were decided upon by the test creators so as not to bias results unfairly towards GPT-4 or OpenAI models.

We found the most effective way to prompt GPT-4 to produce a nuanced emotional scene was to specify that it must depict a scene of conflict or tension. Without this designation, it would typically generate dialogue that lacked emotional nuance and which resolved predictably. As such we elected to narrow the scope to include only scenes of conflict and tension (in a positive or negative context). We seeded each prompt with a random selection of location, author style and a broad scenario description to promote diversity of ideas. The resulting dialogues were varied and creative, and we expect that they will assess a sufficiently broad range of capabilities as to be representative of overall EU.

\subsection{Questions and Reference Answers}

All questions and reference answers were determined by the authors of this paper. The four candidate emotions were selected with the intent of revealing a wide range of EU. A selection of four possible emotions were chosen: Some clearly unlikely, some clearly plausible, and some that require a careful reading and strong EU to delineate. Reference answers were chosen according to the best interpretation of the paper's authors. We acknowledge the limitations of this approach, but point to our results in \autoref{tab:eqbench} to justify this approach as sufficiently effective given resource constraints.

\subsection{The Prompt}

\begin{figure}[ht]
\centering
\begin{quoting}[leftmargin=0.5cm, rightmargin=0.5cm]
\begin{mdframed}
Your task is to predict the likely emotional responses of a character in this dialogue:
\newline

Cecilia: You know, your words have power, Brandon. More than you might think.

Brandon: I'm well aware, Cecilia. It's a critic's job to wield them.

Cecilia: But do you understand the weight of them? The lives they can shatter?

Brandon: Art is not for the faint-hearted. If you can't handle the critique, you're in the wrong industry.

Cecilia: It's not about handling criticism, Brandon. It's about understanding the soul of the art. You dissect it like a cold, lifeless body on an autopsy table.
\newline

[End dialogue]
\newline

At the end of this dialogue, Brandon would feel...
\newline

Offended

Empathetic

Confident

Dismissive
\newline

Give each of these possible emotions a score from 0-10 for the relative intensity that they are likely to be feeling each. Then critique your answer by thinking it through step by step. Finally, give your revised scores.
\end{mdframed}
\end{quoting}
\caption{Example (partial) prompt for a question in the EQ-Bench test set. Further instruction is given for the specific output format, to ensure the answer can be parsed automatically (for the full prompt, see \autoref{app:full_prompt}.}
\label{fig:prompt}
\end{figure}

\subsection{Calculation of Scores}

The scoring process for each question involves several steps:

Normalisation: Firstly, the four emotion intensity ratings are normalised to sum to 10.

Difference Calculation: The sum of the differences between the normalised ratings and the reference (which is already normalised) is calculated. This provides a measure of how closely the model's ratings align with the reference. The score for an individual question is calculated as:

\begin{quoting}[leftmargin=0.5cm, rightmargin=0.5cm]
10 - (sum of differences to reference answers)
\end{quoting}

Subtracting the differences from 10 ensures that a smaller sum of differences (indicating closer alignment with the reference) results in a higher score. The constant of 10 was chosen because it produces an overall score of 0 if questions are answered randomly.

\subsection{Example of Question Scoring}

Consider a subject who provides the following emotional intensity ratings (shown alongside the normalised reference answer):

\begin{center}
\begin{tabular}{ >{\bfseries}l | c | c }
Emotion & Subject Answer & Reference Answer \\
\hline
Offended   & 6 & 1 \\
Empathetic & 0 & 0 \\
Confident  & 7 & 4 \\
Dismissive & 7 & 5 \\
\end{tabular}
\end{center}

We first normalise the subject's ratings to sum to 10, yielding normalised scores of 3, 0, 3.5 and 3.5 respectively. Then we calculate the sum of the absolute differences to the reference answers and subtract from 10:

\begin{align*}
\text{Score} &= 10 - \left( \left| 3-1 \right| + \left| 0-0 \right| + \left| 3.5-4 \right| + \left| 3.5-5 \right| \right) \\
&= 10 - \left( 2 + 0 + 0.5 + 1.5 \right) \\
&= 6
\end{align*}

\subsection{Final Benchmark Score Calculation}

The final score is the \textbf{average of scores for all answers that were parsable.} To focus on assessing emotional intelligence rather than formatting proficiency, only parsable answers are considered in the final score average. In practice, most models produced near to 60 parsable answers. If fewer than 50 out of 60 questions were parsable, the test is considered a fail.

We calculate two final scores: one for the first-pass answer, and one for the revised answers, to allow for models that may perform better at the first pass or the revised pass. This effectively gives the model two runs at the benchmark, taking its best result. We elected not to take the best answer on a per-question basis, as this would have amplified the benefit of making wild guesses, which disproportionately advantages weaker models.

\subsection{Rationale for Normalisation}

Each test question requires the subject to rate the intensity of the four listed emotions from 0-10. Acknowledging the related complexities of interpreting subjective self-ratings of pain in the medical field \citep{karcioglu2018systematic}, we hope to sidestep this potential source variability when scoring answers. To nullify the inherent subjectivity of deciding how intense an emotion ought to be on the 0-10 scale, we normalise both the subject's answers and the reference answers such that all four emotions sum to 10. This normalisation shifts the focus of what is being assessed, such that we are exclusively focusing on the subject's understanding of the \textit{relative intensity of each emotion,} and not caring about the absolute intensity of each rating.

\subsection{Interpretation of Scores}

Multiple choice tests are typically scored by calculating the average number of questions correctly answered. In contrast, EQ-Bench answers are not scored discretely as \textit{wrong} or \textit{right}; instead they are scored according to the distance from the reference answer. The scoring calculation is calibrated such that \textbf{a score of 0 corresponds to answering randomly}. This provides a meaningful baseline.

A score of 100 denotes perfect alignment with the reference answers. However such a score is not realistically achievable given the interpretive nature of the questions. It should be noted that EQ-Bench scores are \textit{not} normalised such that 100 represents the human mean. Negative scores are possible, although in practice, all models tested produced a positive score. We expect that as model performance improves beyond the current state-of-the-art, score variance will be influenced more strongly by differences in interpretation of the questions. 

\subsection{Testing Protocol \& Pipeline}

We developed a testing pipeline in Python, allowing for batch benchmarking of OpenAI models and open-source models in a standardised, automated fashion. Alongside the paper, we will be releasing this code and the test questions under the MIT license. It can be found at \href{https://github.com/EQ-bench/EQ-Bench}{https://github.com/EQ-bench/EQ-Bench}, along with the prompts we used.

Questions are administered in a zero-shot format (i.e. no example inputs/outputs are provided) to minimise any biasing effect on the model's answers. In a similar vein to "think step-by-step" prompting, we ask the model to critique its original answer, and then give a revised answer. This allows the model to interact with its prior reasoning when producing its final answer, a technique that has been demonstrated to improve reasoning with zero-shot prompting \citep{kojima2022large}.

\textbf{To ensure fair comparisons of EQ-Bench scores, we request that users of the benchmark also use zero-shot prompting as we have done in this paper, using the exact prompts and scoring protocol defined here and in our github repository.}

To minimise variance between benchmark runs, inference was generated with a temperature parameter of 0.01. If the answer could not be parsed, the temperature parameter was increased by 0.15 progressively until either a parseable answer was produced or the number of attempts exceeded 5. For the open-source models tested, models were quantised to fit into the available VRAM using the bitsandbytes library \citep{dettmers20218}: 8-bit quantisation for models with 7B - 34B parameters, and 4-bit quantisation for larger models. We acknowledge that quantising may have reduced the score for the models tested, and we hope to quantify this potential effect in future work.

\newpage

\section{Results}

\subsection{EQ-Bench Scores}

\begin{table}[htbp]
\centering
\caption{EQ-Bench Score Comparison}
\label{tab:eqbench}
\begin{tabular}{@{}lr@{}}
\toprule
Model                                   & EQ-Bench Score \\ \midrule
OpenAI gpt-4-0613                       & 62.52          \\
migtissera/SynthIA-70B-v1.5	           & 54.83          \\
OpenAI gpt-4-0314                       & 53.39          \\
Qwen/Qwen-72B-Chat                      & 52.44          \\
Anthropic Claude2                       & 52.14          \\
meta-llama/Llama-2-70b-chat-hf          & 51.56          \\
01-ai/Yi-34B-Chat                       & 51.03          \\
OpenAI gpt-3.5-0613                     & 49.17          \\
OpenAI gpt-3.5-turbo-0301               & 47.61          \\
Open-Orca/Mistral-7B-OpenOrca           & 44.40          \\
Qwen/Qwen-14B-Chat                      & 43.76          \\
OpenAI text-davinci-003                 & 43.73          \\
Intel/neural-chat-7b-v3-1               & 43.61          \\
OpenAI text-davinci-002                 & 39.44          \\
openchat/openchat\_3.5                  & 37.08          \\
lmsys/vicuna-33b-v1.3                   & 36.52          \\
meta-llama/Llama-2-13b-chat-hf          & 33.02          \\
lmsys/vicuna-13b-v1.1                   & 32.85          \\
meta-llama/Llama-2-7b-chat-hf           & 25.43          \\
Koala 13B                               & 24.92          \\
lmsys/vicuna-7b-v1.1                    & 22.24          \\
OpenAI text-davinci-001                 & 15.19          \\
OpenAI Curie                            & 9.28           \\
OpenAI ADA                              & 2.25           \\
OpenAI Babbage                          & FAIL           \\ \bottomrule
\end{tabular}
\end{table}

OpenAI's GPT-4-0613 model produced the highest EQ-Bench score by a considerable margin. We consider this to match with the community consensus of it being the strongest model currently available. Notably, open-source models are rapidly closing the gap with the state-of-the-art proprietary models. The highest performing open-source model tested was SynthIA-70B, a fine-tuned version of Llama2-70B made for role-play. It is interesting to consider that fine-tuning a model specifically for role-play may be an effective way to increase emotional intelligence.

Score differences between different generations of GPT-3.5 and GPT-4 models are evident, with newer versions performing better. Older open-source models such as Koala 13B and Vicuna 7B scored near the bottom of the list.

All models listed in \autoref{tab:eqbench} produced enough parseable answers (at least 50) for a passing score, with the exception of OpenAI's older generation Curie, ADA and Babbage models. By simplifying the prompt to remove the critique \& revision section, we were also able to get a passing score from Curie and ADA, however these scores are close to zero, indicating that their answers were nearly indistinguishable from answering randomly.

\newpage

\subsection{Repeatability}

The benchmark demonstrated good repeatability over multiple benchmark runs for the models tested, with an average 2.93\% CV (see \autoref{app:repeatability} for individual scores). This variance is due to the non-deterministic nature of LLM output. In practice, this variance may be reduced by benchmarking a model multiple times and averaging the result, which can be done simply in our benchmarking pipeline.

\subsection{Effect of Critique \& Score Revision}

\begin{figure}[htbp]
\centering
\includegraphics[width=400px]{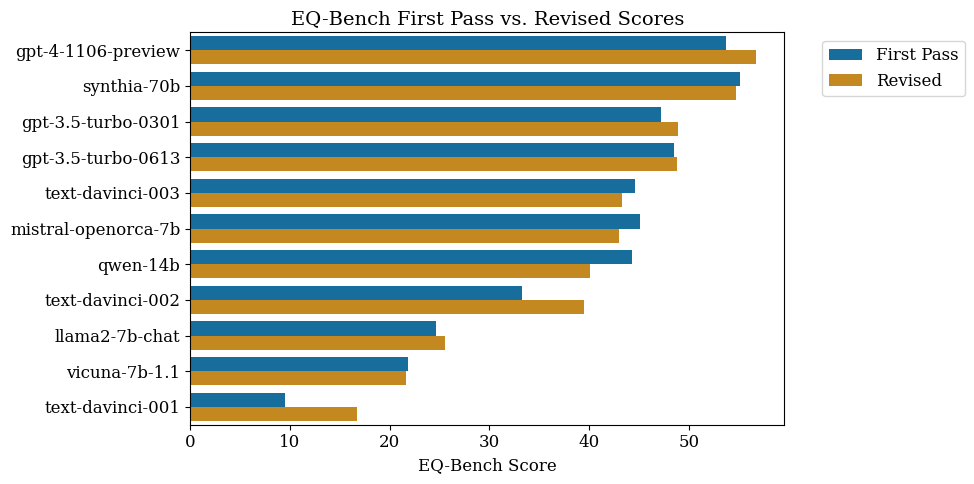}
\caption{EQ-Bench scores comparing first-pass scores vs. revised scores.}
\label{fig:revisedscores}
\end{figure}

The method of prompting models to critique and revise their answers \textbf{improved scores on average by 9.3\%} amongst the models compared (see \autoref{fig:revisedscores}). The benefit of revision was inconsistent: lower performing models showed a more pronounced benefit, and some models resulted in a decreased score for their revised answers.

We include this technique in our benchmark protocol primarily to give the model a chance to deploy its best reasoning capabilities. Benchmarkers will often modify a benchmark protocol to incorporate multi-shot prompting or chain-of-thought reasoning, in order to achieve higher scores. By releasing our benchmarking pipeline code and defining the prompting protocol explicitly, we hope to mitigate some potential sources of variance when the benchmark is used in the wild.

\newpage

\subsection{Comparing SECEU EQ vs. EQ-Bench Scores}

\begin{figure}[htbp]
\centering
\includegraphics[width=400px]{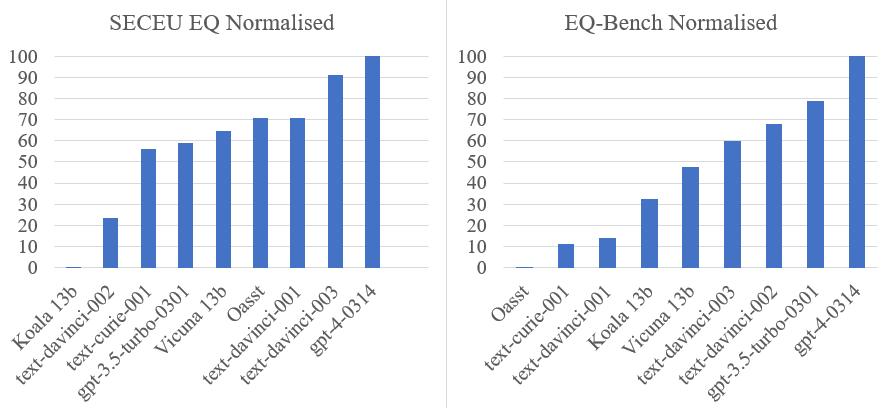}
\caption{Visualisation of Score Distributions (normalised 0-100) for SECEU EQ vs. EQ-Bench (see \autoref{app:score_comparison} for table of scores).}
\label{fig:normscores}
\end{figure}

Of immediate note, there is significant disagreement between EQ-Bench scores and SECEU EQ scores. While it is difficult to determine from the scores alone why this disagreement might exist, we can normalise the scores to compare them side-by-side visually (\autoref{fig:normscores}) and examine the distributions statistically.

\begin{table}[h]
\centering
\caption{Calculated Statistics for Normalised Benchmark Scores}
\begin{tabular}{l>{\raggedright\arraybackslash}p{1.8cm}>{\raggedright\arraybackslash}p{1.8cm}>{\raggedright\arraybackslash}p{1.8cm}>{\raggedright\arraybackslash}p{1.8cm}}
\hline
\textbf{Metric} & \textbf{SECEU EQ} & \textbf{EQ-Bench (Ours)} \\
\hline
Mean                      & 59.48 & 45.75 \\
Median                      & 64.71 & 47.44 \\
Coefficient of Variation (CV) & 0.52 & 0.74 \\
Interquartile Range (IQR)     & 14.71 & 53.81 \\
Skewness                      & -0.68 & 0.13 \\
\hline
\end{tabular}
\label{tab:stats}
\end{table}

\noindent \autoref{tab:stats} presents a comparison of statistical measures calculated for SECEU EQ and EQ-Bench score distributions. To produce a fair comparison, we include only the models in the SECEU paper that we were able to successfully benchmark with EQ-Bench. Each series of scores was normalised to 0-100.

Visually it is evident that SECEU EQ scores exhibit more bunching around its median of 64.71. This is borne out by the relatively low Inter-quartile range (IQR) of 14.71 compared to EQ-Bench's IQR of 53.81. The Coefficient of Variation (CV) and Skewness values also demonstrate a wider, more symmetrical spread for EQ-Bench scores.

It must be noted that these statistical comparisons of score distribution do not take into account whether the scores are accurately measuring emotional understanding. To gain further insight into that aspect of performance, we can look at the correlation of EQ-Bench scores with other popular benchmarks.

\newpage

\subsection{Correlation with Other Benchmarks}

\begin{figure}[htbp]
\centering
\includegraphics[width=\textwidth]{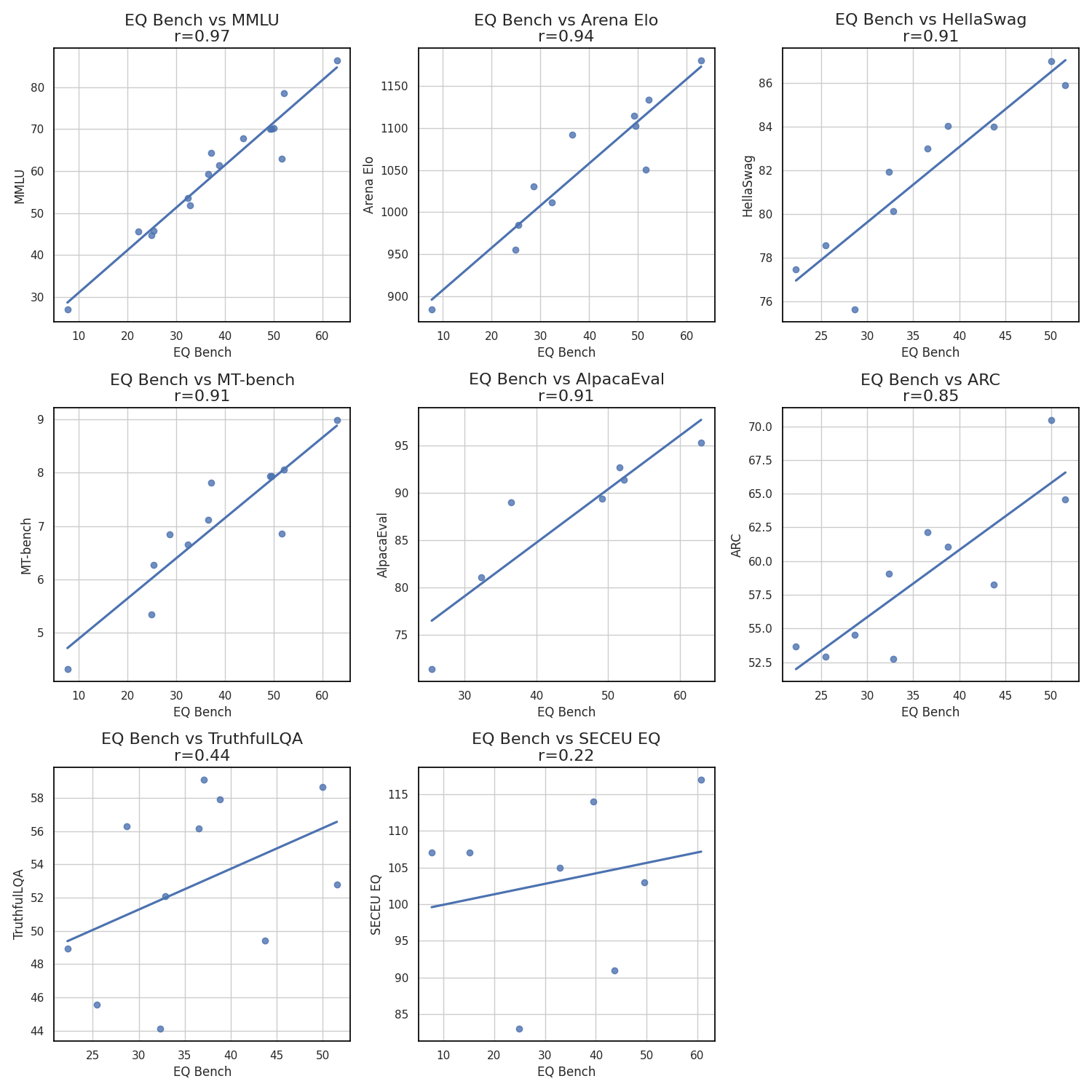}
\caption{Scatter plots of EQ-Bench scores vs. other popular benchmarks, showing strong correlation with MMLU \citep{hendrycks2020measuring}, HellaSwag \citep{zellers2019hellaswag}, AlpacaEval \citep{li2023alpacaeval}, Chatbot Arena ELO \citep{zheng2023judging}, ARC \citep{clark2018think} and MT-bench \citep{zheng2023judging}. Weaker correlations were observed with TruthfulQA \citep{lin2021truthfulqa} and SECEU EQ \citep{wang2023emotional}. For the full score correlation matrix see \autoref{app:score_matrix1} and \autoref{app:score_matrix2}.}
\label{fig:scatter}
\end{figure}

To examine the agreement of EQ-Bench with other established benchmarks, we computed Pearson correlation coefficients and scatter plots of scores (see Figure \ref{fig:scatter}).

\newpage

\section{Discussion}

A criticism often levelled at LLM benchmarks is that high scores on the benchmark do not necessarily correlate to real world performance. There are frequent demonstrations of open-source models beating the likes of GPT-4 in specific benchmarks, while not being nearly as capable in regular use. We perceive this to be a source of significant distrust of synthetic benchmark scores as a way of judging a model's capabilities. This is a primary motivation for creating EQ-Bench: to produce benchmark scores that reliably correlate with the user experience and perceptions of the intelligence of the model.

\subsection{Correlation with Existing Multi-Domain Benchmarks}

Our results correlate most strongly with MMLU (r=0.97), which is an industry standard benchmark of some 14 thousand questions covering a wide range of knowledge domains and capabilities. EQ-Bench also correlates strongly with other multi-domain benchmarks of this type: HellaSwag (r=0.91) and ARC (0.85). We take this as convincing evidence that our approach of using a narrow set of questions targeting EU produces generalisable results that are representative of a model's broad intelligence.

Our interpretation of these strong correlations is that measuring EU in this way inherently tests complex features of cognition, understanding and reasoning, and thus has significant overlap with the broad features of intelligence tested by multi-domain benchmarks. The strength of these correlations supports our hypothesis that EU measurements can act as a proxy for broad intelligence in LLMs.

\subsection{Correlation with Perceived Intelligence of the Model}

EQ-Bench scores show a strong Pearson correlation (r=0.94) with LMSYS's Chatbot Arena ELO scores \citep{zheng2023judging}, which are derived from human preferences in head-to-head comparisons of model outputs. EQ-Bench scores also align closely with benchmarks that use GPT-4 as a judge of the quality of model output: AlpacaEval \citep{li2023alpacaeval}: r=0.91, and MT-Bench \citep{zheng2023judging}: r=0.91. These benchmarks are not measuring precisely the same thing as \textit{the perceived intelligence of the model}, but we think the underlying thing being measured is similar, and the close correlations of scores is likely indicative.

We find anecdotally that EQ-Bench scores align closely with our perception of the relative intelligence and capabilities of the models tested. We observe scores to typically be more sensitive to parameter size differences of models when compared to other benchmarks. For example, comparing a 7 billion parameter vs. a 70 billion parameter model:

\begin{itemize}
\item Meta's 70b Llama2 chat model \citep{touvron2023llama} scored only \textbf{1.4\% higher} than the OpenOrca fine-tune of the Mistral-7b foundational model on an aggregate of four popular benchmarks: ARC, HellaSwag, MMLU and TruthfulQA \citep{lian2023mistralorca}.
\item Whereas EQ-Bench scored the 70 billion parameter Llama2 model \textbf{13.9\% higher} than the 7 billion parameter Mistral-based model.
\end{itemize}

In this case we find that our result better captures the relative performance disparity between the two models than the other cited benchmarks. We highlight this as an example of a fine-tuned model perhaps overfitting the kinds of questions found in these benchmarks, resulting in a disproportionate gain in scores relative to the perceived holistic improvement in model performance and the disparity in model parameter counts.

\newpage

\subsection{Comparison with SECEU}

EQ-Bench produces results that notably correlate weakly with SECEU, despite the similarity of the question format and the fact that we are both specifically targeting EU. We believe the discrepancy is explained by the methodological improvements we implemented with EQ-Bench. Specifically:

\begin{itemize}
\item Dialogues focused on nuanced emotional interactions.
\item A more diverse selection of emotions to rate (as opposed to the most plausible ones).
\item Choosing reference answers ourselves rather than crowdsourcing.
\item Removing the requirement to sum intensity ratings to 10.
\end{itemize}

\begin{figure}[htbp]
\centering
\includegraphics[width=\textwidth]{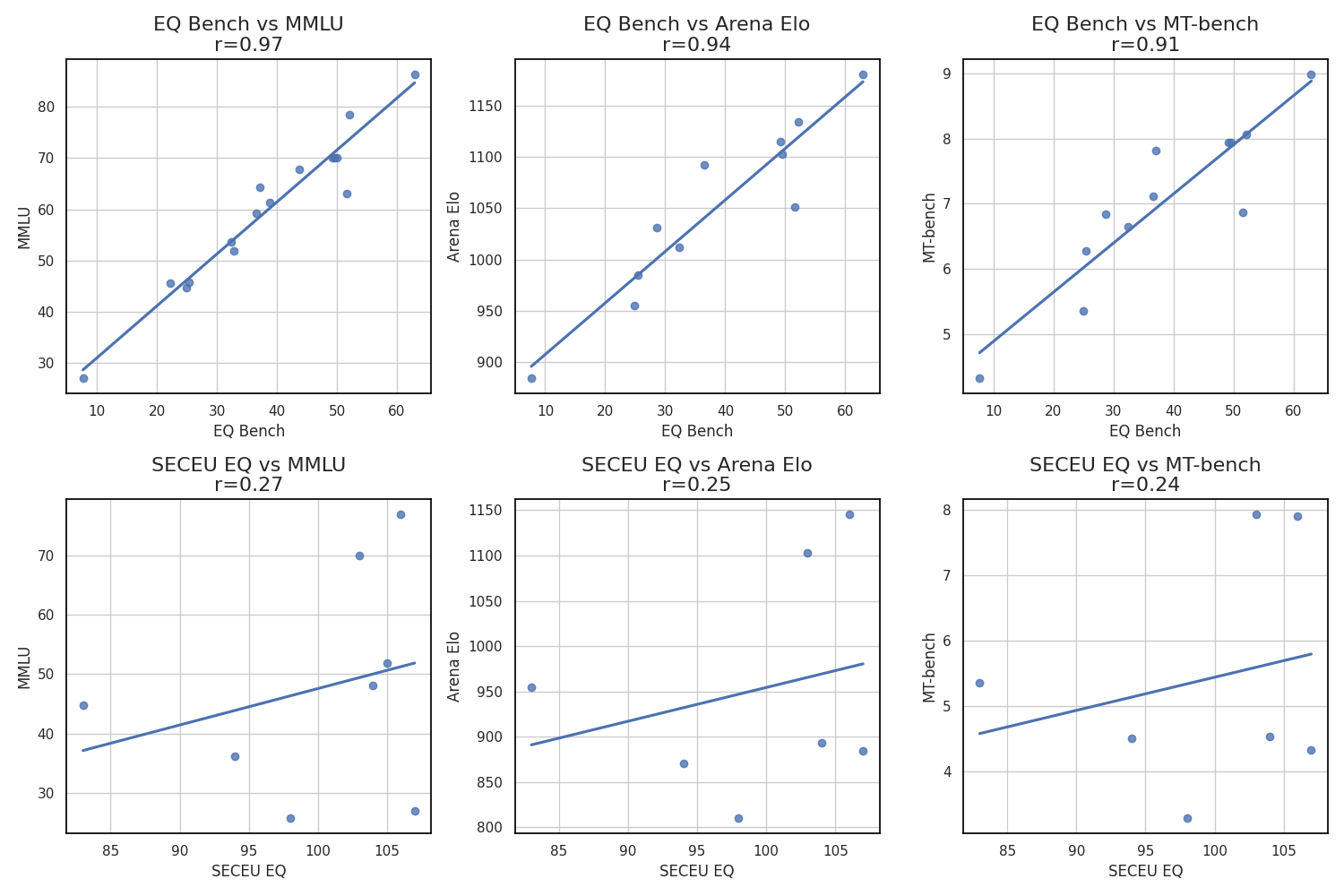}
\caption{Comparison of correlations between EQ-Bench and SECEU vs. other benchmarks. EQ-Bench produces scores that correlate much more strongly with industry standard benchmarks compared to SECEU. Data for the scores and models being compared can be found in \autoref{app:score_matrix1} and \autoref{app:score_matrix2}.}
\label{fig:scatter_seceu}
\end{figure}

We have shown strong agreement with industry standard benchmarks (\autoref{fig:scatter}), and strong statistical differentiation of scores compared to SECEU as outlined in \autoref{tab:stats}. EQ-Bench produces scores that correlate much more strongly with industry standard benchmarks (see \autoref{fig:scatter_seceu}).

\subsection{Difficult to "Game" or Cheat the Benchmark}

While we have not attempted to adversarially fine-tune a model to score well in the benchmark, we anticipate that it will not be trivial to do this without outright cheating by training on the test questions. This is due to several factors:

\textbf{Complex nature of EU:} Emotional understanding represents a set of complex cognitive abilities. We expect it will require significant engineering and curation of training sets in order to increase a model's EU through fine tuning. In contrast, other benchmarks may be more susceptible to narrow improvements in a model fine-tuned specifically to improve benchmark scores.

\textbf{Objective scoring:} EQ-Bench is scored without the need of subjective interpretation by assessors. This eliminates potential sources of bias that might be exploited in LLM-as-a-judge benchmarks.

\textbf{Training set leaks:} Benchmark questions can leak into the training set of a model, artificially inflating its scores. This is a risk for any benchmark, however it may be detectable after the fact, and this is something we intend to explore in future research.

We acknowledge that in practice, LLM benchmark leaderboards are highly competitive and methods may be found to artificially inflate a model's score on EQ-Bench. As such we will continue to monitor the evolving space and consider adjustments to our methodology to improve its robustness.

\subsection{Able to Differentiate a Wide Range of EU Levels}

We observe that scores are well distributed through the whole range of models tested, without apparent bunching or levelling off. The Coefficient of Variation (0.741), Interquartile Range (53.81) and Skewness (0.13) of normalised EQ-Bench scores indicate a broad spread and near-symmetrical distribution. The lack of evident bunching of scores (see \autoref{tab:eqbench}) suggests that the ceiling for the test's ability to measure EU is at least as high as the strongest models tested. It remains to be seen how much headroom there is to differentiate models above this.

\subsection{Runtime and Computational Requirements}

In our testing, an EQ-Bench run typically completes in less than 10 mins for OpenAI models. For open source models, the test typically completes in 20-60 minutes when running on a single Nvidia RTX A6000. These requirements make the test accessible and economical compared to other benchmarks that require thousands of questions to be answered. We are releasing a python pipeline on Github to streamline the process of benchmarking multiple models in a single run without human intervention.

\section{Limitations and Future Directions}

The inherent subjectivity in predicting emotional responses means that there is no objectively correct answer to the test questions. Efforts have been made to minimize ambiguity by selecting dialogues and emotions whose intensity can be delineated without ambiguity. We recognise that these are inherent concerns in the design of any psychometric test, and while ultimately unavoidable, we may in future be able to mitigate such ambiguities by designing questions in consultation with domain experts.

Likewise, the range of emotional understanding that the test is able to measure is inherently limited to the abilities of the authors of this paper to create challenging questions and to set the reference answers insightfully. We expect scores to compress as model capabilities approach the upper limit that the test is able to measure. In order to increase the ceiling of EU that the benchmark is able to measure, we may in future work employ experts in EI to craft more complex questions and to collectively decide on the reference answers.

While the benchmark reference answers were determined by the authors of this paper, all dialogues were generated by GPT-4. The benchmark could be improved by employing human writers to create the dialogues, eliminating a potential source of bias and improving the emotional depth and complexity of the scenes.

Due to resource constraints, we were not able to administer the test on a cohort of human subjects in order to establish this point of reference. This may be worthwhile for future work, since normalising scores to 100 at the human mean allows for an intuitive point of reference with respect to other psychometrics like IQ. It would also be valuable to administer industry standard EI tests on several language models with the help of accredited assessors. These results could be correlated with EQ-Bench scores, to determine if we are effectively measuring similar aspects of EI.

\newpage

\bibliographystyle{apacite}
\bibliography{references}

\begin{thebibliography}{}

\bibitem [\protect \citeauthoryear {%
Clark%
\ \protect \BOthers {.}}{%
Clark%
\ \protect \BOthers {.}}{%
{\protect \APACyear {2018}}%
}]{%
clark2018think}
\APACinsertmetastar {%
clark2018think}%
\begin{APACrefauthors}%
Clark, P.%
, Cowhey, I.%
, Etzioni, O.%
, Khot, T.%
, Sabharwal, A.%
, Schoenick, C.%
\BCBL {}\ \BBA {} Tafjord, O.%
\end{APACrefauthors}%
\unskip\
\newblock
\APACrefYearMonthDay{2018}{}{}.
\newblock
{\BBOQ}\APACrefatitle {Think you have solved question answering? try arc, the
  ai2 reasoning challenge} {Think you have solved question answering? try arc,
  the ai2 reasoning challenge}.{\BBCQ}
\newblock
\APACjournalVolNumPages{arXiv preprint arXiv:1803.05457}{}{}{}.
\PrintBackRefs{\CurrentBib}

\bibitem [\protect \citeauthoryear {%
Dettmers%
, Lewis%
, Shleifer%
\BCBL {}\ \BBA {} Zettlemoyer%
}{%
Dettmers%
\ \protect \BOthers {.}}{%
{\protect \APACyear {2021}}%
}]{%
dettmers20218}
\APACinsertmetastar {%
dettmers20218}%
\begin{APACrefauthors}%
Dettmers, T.%
, Lewis, M.%
, Shleifer, S.%
\BCBL {}\ \BBA {} Zettlemoyer, L.%
\end{APACrefauthors}%
\unskip\
\newblock
\APACrefYearMonthDay{2021}{}{}.
\newblock
{\BBOQ}\APACrefatitle {8-bit optimizers via block-wise quantization} {8-bit
  optimizers via block-wise quantization}.{\BBCQ}
\newblock
\APACjournalVolNumPages{arXiv preprint arXiv:2110.02861}{}{}{}.
\PrintBackRefs{\CurrentBib}

\bibitem [\protect \citeauthoryear {%
Elyoseph%
, Hadar-Shoval%
, Asraf%
\BCBL {}\ \BBA {} Lvovsky%
}{%
Elyoseph%
\ \protect \BOthers {.}}{%
{\protect \APACyear {2023}}%
}]{%
elyoseph2023chatgpt}
\APACinsertmetastar {%
elyoseph2023chatgpt}%
\begin{APACrefauthors}%
Elyoseph, Z.%
, Hadar-Shoval, D.%
, Asraf, K.%
\BCBL {}\ \BBA {} Lvovsky, M.%
\end{APACrefauthors}%
\unskip\
\newblock
\APACrefYearMonthDay{2023}{}{}.
\newblock
{\BBOQ}\APACrefatitle {ChatGPT outperforms humans in emotional awareness
  evaluations} {Chatgpt outperforms humans in emotional awareness
  evaluations}.{\BBCQ}
\newblock
\APACjournalVolNumPages{Frontiers in Psychology}{14}{}{1199058}.
\PrintBackRefs{\CurrentBib}

\bibitem [\protect \citeauthoryear {%
Goleman%
}{%
Goleman%
}{%
{\protect \APACyear {1996}}%
}]{%
goleman1996emotional}
\APACinsertmetastar {%
goleman1996emotional}%
\begin{APACrefauthors}%
Goleman, D.%
\end{APACrefauthors}%
\unskip\
\newblock
\APACrefYearMonthDay{1996}{}{}.
\newblock
{\BBOQ}\APACrefatitle {Emotional intelligence. Why it can matter more than IQ.}
  {Emotional intelligence. why it can matter more than iq.}{\BBCQ}
\newblock
\APACjournalVolNumPages{Learning}{24}{6}{49--50}.
\PrintBackRefs{\CurrentBib}

\bibitem [\protect \citeauthoryear {%
Hendrycks%
\ \protect \BOthers {.}}{%
Hendrycks%
\ \protect \BOthers {.}}{%
{\protect \APACyear {2020}}%
}]{%
hendrycks2020measuring}
\APACinsertmetastar {%
hendrycks2020measuring}%
\begin{APACrefauthors}%
Hendrycks, D.%
, Burns, C.%
, Basart, S.%
, Zou, A.%
, Mazeika, M.%
, Song, D.%
\BCBL {}\ \BBA {} Steinhardt, J.%
\end{APACrefauthors}%
\unskip\
\newblock
\APACrefYearMonthDay{2020}{}{}.
\newblock
{\BBOQ}\APACrefatitle {Measuring massive multitask language understanding}
  {Measuring massive multitask language understanding}.{\BBCQ}
\newblock
\APACjournalVolNumPages{arXiv preprint arXiv:2009.03300}{}{}{}.
\PrintBackRefs{\CurrentBib}

\bibitem [\protect \citeauthoryear {%
Hendrycks%
\ \protect \BOthers {.}}{%
Hendrycks%
\ \protect \BOthers {.}}{%
{\protect \APACyear {2021}}%
}]{%
hendrycks2021measuring}
\APACinsertmetastar {%
hendrycks2021measuring}%
\begin{APACrefauthors}%
Hendrycks, D.%
, Burns, C.%
, Kadavath, S.%
, Arora, A.%
, Basart, S.%
, Tang, E.%
\BDBL {}Steinhardt, J.%
\end{APACrefauthors}%
\unskip\
\newblock
\APACrefYearMonthDay{2021}{}{}.
\newblock
{\BBOQ}\APACrefatitle {Measuring mathematical problem solving with the math
  dataset} {Measuring mathematical problem solving with the math
  dataset}.{\BBCQ}
\newblock
\APACjournalVolNumPages{arXiv preprint arXiv:2103.03874}{}{}{}.
\PrintBackRefs{\CurrentBib}

\bibitem [\protect \citeauthoryear {%
Huyghe%
, Hovasapian%
\BCBL {}\ \BBA {} Fontaine%
}{%
Huyghe%
\ \protect \BOthers {.}}{%
{\protect \APACyear {2022}}%
}]{%
huyghe2022scoring}
\APACinsertmetastar {%
huyghe2022scoring}%
\begin{APACrefauthors}%
Huyghe, V\BPBI E.%
, Hovasapian, A.%
\BCBL {}\ \BBA {} Fontaine, J\BPBI R.%
\end{APACrefauthors}%
\unskip\
\newblock
\APACrefYearMonthDay{2022}{}{}.
\newblock
{\BBOQ}\APACrefatitle {The scoring challenge of Emotional Intelligence ability
  tests: A Confirmatory Factor Analysis approach to model substantive and
  method effects using raw item scores} {The scoring challenge of emotional
  intelligence ability tests: A confirmatory factor analysis approach to model
  substantive and method effects using raw item scores}.{\BBCQ}
\newblock
\APACjournalVolNumPages{Frontiers in Psychology}{13}{}{812525}.
\PrintBackRefs{\CurrentBib}

\bibitem [\protect \citeauthoryear {%
Karcioglu%
, Topacoglu%
, Dikme%
\BCBL {}\ \BBA {} Dikme%
}{%
Karcioglu%
\ \protect \BOthers {.}}{%
{\protect \APACyear {2018}}%
}]{%
karcioglu2018systematic}
\APACinsertmetastar {%
karcioglu2018systematic}%
\begin{APACrefauthors}%
Karcioglu, O.%
, Topacoglu, H.%
, Dikme, O.%
\BCBL {}\ \BBA {} Dikme, O.%
\end{APACrefauthors}%
\unskip\
\newblock
\APACrefYearMonthDay{2018}{}{}.
\newblock
{\BBOQ}\APACrefatitle {A systematic review of the pain scales in adults: which
  to use?} {A systematic review of the pain scales in adults: which to
  use?}{\BBCQ}
\newblock
\APACjournalVolNumPages{The American journal of emergency
  medicine}{36}{4}{707--714}.
\PrintBackRefs{\CurrentBib}

\bibitem [\protect \citeauthoryear {%
Kojima%
, Gu%
, Reid%
, Matsuo%
\BCBL {}\ \BBA {} Iwasawa%
}{%
Kojima%
\ \protect \BOthers {.}}{%
{\protect \APACyear {2022}}%
}]{%
kojima2022large}
\APACinsertmetastar {%
kojima2022large}%
\begin{APACrefauthors}%
Kojima, T.%
, Gu, S\BPBI S.%
, Reid, M.%
, Matsuo, Y.%
\BCBL {}\ \BBA {} Iwasawa, Y.%
\end{APACrefauthors}%
\unskip\
\newblock
\APACrefYearMonthDay{2022}{}{}.
\newblock
{\BBOQ}\APACrefatitle {Large language models are zero-shot reasoners} {Large
  language models are zero-shot reasoners}.{\BBCQ}
\newblock
\APACjournalVolNumPages{Advances in neural information processing
  systems}{35}{}{22199--22213}.
\PrintBackRefs{\CurrentBib}

\bibitem [\protect \citeauthoryear {%
Li%
\ \protect \BOthers {.}}{%
Li%
\ \protect \BOthers {.}}{%
{\protect \APACyear {2023}}%
}]{%
li2023alpacaeval}
\APACinsertmetastar {%
li2023alpacaeval}%
\begin{APACrefauthors}%
Li, X.%
, Zhang, T.%
, Dubois, Y.%
, Taori, R.%
, Gulrajani, I.%
, Guestrin, C.%
\BDBL {}Hashimoto, T\BPBI B.%
\end{APACrefauthors}%
\unskip\
\newblock
\APACrefYearMonthDay{2023}{}{}.
\newblock
{\BBOQ}\APACrefatitle {Alpacaeval: An automatic evaluator of
  instruction-following models} {Alpacaeval: An automatic evaluator of
  instruction-following models}.{\BBCQ}
\newblock
\APACjournalVolNumPages{GitHub repository}{}{}{}.
\PrintBackRefs{\CurrentBib}

\bibitem [\protect \citeauthoryear {%
Lian%
\ \protect \BOthers {.}}{%
Lian%
\ \protect \BOthers {.}}{%
{\protect \APACyear {2023}}%
}]{%
lian2023mistralorca}
\APACinsertmetastar {%
lian2023mistralorca}%
\begin{APACrefauthors}%
Lian, W.%
, Goodson, B.%
, Wang, G.%
, Pentland, E.%
, Cook, A.%
, Vong, C.%
\BCBL {}\ \BBA {} "Teknium".%
\end{APACrefauthors}%
\unskip\
\newblock
\APACrefYearMonthDay{2023}{}{}.
\newblock
\APACrefbtitle {MistralOrca: Mistral-7B Model Instruct-tuned on Filtered
  OpenOrcaV1 GPT-4 Dataset.} {Mistralorca: Mistral-7b model instruct-tuned on
  filtered openorcav1 gpt-4 dataset.}
\newblock
\APAChowpublished {\url{https://huggingface.co/Open-Orca/Mistral-7B-OpenOrca}}.
\newblock
\APACaddressPublisher{}{HuggingFace}.
\PrintBackRefs{\CurrentBib}

\bibitem [\protect \citeauthoryear {%
Lin%
, Hilton%
\BCBL {}\ \BBA {} Evans%
}{%
Lin%
\ \protect \BOthers {.}}{%
{\protect \APACyear {2021}}%
}]{%
lin2021truthfulqa}
\APACinsertmetastar {%
lin2021truthfulqa}%
\begin{APACrefauthors}%
Lin, S.%
, Hilton, J.%
\BCBL {}\ \BBA {} Evans, O.%
\end{APACrefauthors}%
\unskip\
\newblock
\APACrefYearMonthDay{2021}{}{}.
\newblock
{\BBOQ}\APACrefatitle {Truthfulqa: Measuring how models mimic human falsehoods}
  {Truthfulqa: Measuring how models mimic human falsehoods}.{\BBCQ}
\newblock
\APACjournalVolNumPages{arXiv preprint arXiv:2109.07958}{}{}{}.
\PrintBackRefs{\CurrentBib}

\bibitem [\protect \citeauthoryear {%
LMSYS%
}{%
LMSYS%
}{%
{\protect \APACyear {2023}}%
}]{%
chatbot_arena_leaderboard}
\APACinsertmetastar {%
chatbot_arena_leaderboard}%
\begin{APACrefauthors}%
LMSYS.%
\end{APACrefauthors}%
\unskip\
\newblock
\APACrefYearMonthDay{2023}{}{}.
\newblock
\APACrefbtitle {Chatbot Arena Leaderboard.} {Chatbot arena leaderboard.}
\newblock
\begin{APACrefURL}
  \url{https://huggingface.co/spaces/lmsys/chatbot-arena-leaderboard}
  \end{APACrefURL}
\newblock
\APACrefnote{Accessed: 2023-12-06}
\PrintBackRefs{\CurrentBib}

\bibitem [\protect \citeauthoryear {%
Mayer%
\ \BBA {} Salovey%
}{%
Mayer%
\ \BBA {} Salovey%
}{%
{\protect \APACyear {1997}}%
}]{%
mayer1997emotional}
\APACinsertmetastar {%
mayer1997emotional}%
\begin{APACrefauthors}%
Mayer, J.%
\BCBT {}\ \BBA {} Salovey, P.%
\end{APACrefauthors}%
\unskip\
\newblock
\APACrefYearMonthDay{1997}{}{}.
\newblock
{\BBOQ}\APACrefatitle {What Is The Emotional Intelligence? Implications for
  Education} {What is the emotional intelligence? implications for
  education}.{\BBCQ}
\newblock
\APACjournalVolNumPages{Emotional Development, Emotional Literacy, and
  Emotional Intelligence, New york: Basic books}{}{}{}.
\PrintBackRefs{\CurrentBib}

\bibitem [\protect \citeauthoryear {%
McRae%
\ \protect \BOthers {.}}{%
McRae%
\ \protect \BOthers {.}}{%
{\protect \APACyear {2017}}%
}]{%
mcrae2017genetic}
\APACinsertmetastar {%
mcrae2017genetic}%
\begin{APACrefauthors}%
McRae, K.%
, Rhee, S\BPBI H.%
, Gatt, J\BPBI M.%
, Godinez, D.%
, Williams, L\BPBI M.%
\BCBL {}\ \BBA {} Gross, J\BPBI J.%
\end{APACrefauthors}%
\unskip\
\newblock
\APACrefYearMonthDay{2017}{}{}.
\newblock
{\BBOQ}\APACrefatitle {Genetic and environmental influences on emotion
  regulation: A twin study of cognitive reappraisal and expressive
  suppression.} {Genetic and environmental influences on emotion regulation: A
  twin study of cognitive reappraisal and expressive suppression.}{\BBCQ}
\newblock
\APACjournalVolNumPages{Emotion}{17}{5}{772}.
\PrintBackRefs{\CurrentBib}

\bibitem [\protect \citeauthoryear {%
Ogurlu%
}{%
Ogurlu%
}{%
{\protect \APACyear {2021}}%
}]{%
ogurlu2021meta}
\APACinsertmetastar {%
ogurlu2021meta}%
\begin{APACrefauthors}%
Ogurlu, U.%
\end{APACrefauthors}%
\unskip\
\newblock
\APACrefYearMonthDay{2021}{}{}.
\newblock
{\BBOQ}\APACrefatitle {A meta-analytic review of emotional intelligence in
  gifted individuals: A multilevel analysis} {A meta-analytic review of
  emotional intelligence in gifted individuals: A multilevel analysis}.{\BBCQ}
\newblock
\APACjournalVolNumPages{Personality and Individual Differences}{171}{}{110503}.
\PrintBackRefs{\CurrentBib}

\bibitem [\protect \citeauthoryear {%
OpenAI%
}{%
OpenAI%
}{%
{\protect \APACyear {2023}}%
}]{%
openai2023gpt4}
\APACinsertmetastar {%
openai2023gpt4}%
\begin{APACrefauthors}%
OpenAI.%
\end{APACrefauthors}%
\unskip\
\newblock
\APACrefYearMonthDay{2023}{}{}.
\newblock
\APACrefbtitle {GPT-4 Technical Report.} {Gpt-4 technical report.}
\PrintBackRefs{\CurrentBib}

\bibitem [\protect \citeauthoryear {%
Salovey%
\ \BBA {} Mayer%
}{%
Salovey%
\ \BBA {} Mayer%
}{%
{\protect \APACyear {1990}}%
}]{%
salovey1990emotional}
\APACinsertmetastar {%
salovey1990emotional}%
\begin{APACrefauthors}%
Salovey, P.%
\BCBT {}\ \BBA {} Mayer, J\BPBI D.%
\end{APACrefauthors}%
\unskip\
\newblock
\APACrefYearMonthDay{1990}{}{}.
\newblock
{\BBOQ}\APACrefatitle {Emotional intelligence} {Emotional intelligence}.{\BBCQ}
\newblock
\APACjournalVolNumPages{Imagination, cognition and
  personality}{9}{3}{185--211}.
\PrintBackRefs{\CurrentBib}

\bibitem [\protect \citeauthoryear {%
Touvron%
\ \protect \BOthers {.}}{%
Touvron%
\ \protect \BOthers {.}}{%
{\protect \APACyear {2023}}%
}]{%
touvron2023llama}
\APACinsertmetastar {%
touvron2023llama}%
\begin{APACrefauthors}%
Touvron, H.%
, Martin, L.%
, Stone, K.%
, Albert, P.%
, Almahairi, A.%
, Babaei, Y.%
\BDBL {}others%
\end{APACrefauthors}%
\unskip\
\newblock
\APACrefYearMonthDay{2023}{}{}.
\newblock
{\BBOQ}\APACrefatitle {Llama 2: Open foundation and fine-tuned chat models}
  {Llama 2: Open foundation and fine-tuned chat models}.{\BBCQ}
\newblock
\APACjournalVolNumPages{arXiv preprint arXiv:2307.09288}{}{}{}.
\PrintBackRefs{\CurrentBib}

\bibitem [\protect \citeauthoryear {%
Turkheimer%
, Haley%
, Waldron%
, d'Onofrio%
\BCBL {}\ \BBA {} Gottesman%
}{%
Turkheimer%
\ \protect \BOthers {.}}{%
{\protect \APACyear {2003}}%
}]{%
turkheimer2003socioeconomic}
\APACinsertmetastar {%
turkheimer2003socioeconomic}%
\begin{APACrefauthors}%
Turkheimer, E.%
, Haley, A.%
, Waldron, M.%
, d'Onofrio, B.%
\BCBL {}\ \BBA {} Gottesman, I\BPBI I.%
\end{APACrefauthors}%
\unskip\
\newblock
\APACrefYearMonthDay{2003}{}{}.
\newblock
{\BBOQ}\APACrefatitle {Socioeconomic status modifies heritability of IQ in
  young children} {Socioeconomic status modifies heritability of iq in young
  children}.{\BBCQ}
\newblock
\APACjournalVolNumPages{Psychological science}{14}{6}{623--628}.
\PrintBackRefs{\CurrentBib}

\bibitem [\protect \citeauthoryear {%
Vaswani%
\ \protect \BOthers {.}}{%
Vaswani%
\ \protect \BOthers {.}}{%
{\protect \APACyear {2017}}%
}]{%
vaswani2017attention}
\APACinsertmetastar {%
vaswani2017attention}%
\begin{APACrefauthors}%
Vaswani, A.%
, Shazeer, N.%
, Parmar, N.%
, Uszkoreit, J.%
, Jones, L.%
, Gomez, A\BPBI N.%
\BDBL {}Polosukhin, I.%
\end{APACrefauthors}%
\unskip\
\newblock
\APACrefYearMonthDay{2017}{}{}.
\newblock
{\BBOQ}\APACrefatitle {Attention is all you need} {Attention is all you
  need}.{\BBCQ}
\newblock
\APACjournalVolNumPages{Advances in neural information processing
  systems}{30}{}{}.
\PrintBackRefs{\CurrentBib}

\bibitem [\protect \citeauthoryear {%
Vernon%
, Petrides%
, Bratko%
\BCBL {}\ \BBA {} Schermer%
}{%
Vernon%
\ \protect \BOthers {.}}{%
{\protect \APACyear {2008}}%
}]{%
vernon2008behavioral}
\APACinsertmetastar {%
vernon2008behavioral}%
\begin{APACrefauthors}%
Vernon, P\BPBI A.%
, Petrides, K.%
, Bratko, D.%
\BCBL {}\ \BBA {} Schermer, J\BPBI A.%
\end{APACrefauthors}%
\unskip\
\newblock
\APACrefYearMonthDay{2008}{}{}.
\newblock
{\BBOQ}\APACrefatitle {A behavioral genetic study of trait emotional
  intelligence.} {A behavioral genetic study of trait emotional
  intelligence.}{\BBCQ}
\newblock
\APACjournalVolNumPages{Emotion}{8}{5}{635}.
\PrintBackRefs{\CurrentBib}

\bibitem [\protect \citeauthoryear {%
Wang%
, Li%
, Yin%
, Wu%
\BCBL {}\ \BBA {} Liu%
}{%
Wang%
\ \protect \BOthers {.}}{%
{\protect \APACyear {2023}}%
}]{%
wang2023emotional}
\APACinsertmetastar {%
wang2023emotional}%
\begin{APACrefauthors}%
Wang, X.%
, Li, X.%
, Yin, Z.%
, Wu, Y.%
\BCBL {}\ \BBA {} Liu, J.%
\end{APACrefauthors}%
\unskip\
\newblock
\APACrefYearMonthDay{2023}{}{}.
\newblock
{\BBOQ}\APACrefatitle {Emotional intelligence of large language models}
  {Emotional intelligence of large language models}.{\BBCQ}
\newblock
\APACjournalVolNumPages{Journal of Pacific Rim
  Psychology}{17}{}{18344909231213958}.
\PrintBackRefs{\CurrentBib}

\bibitem [\protect \citeauthoryear {%
Yang%
\ \protect \BOthers {.}}{%
Yang%
\ \protect \BOthers {.}}{%
{\protect \APACyear {2023}}%
}]{%
yang2023dawn}
\APACinsertmetastar {%
yang2023dawn}%
\begin{APACrefauthors}%
Yang, Z.%
, Li, L.%
, Lin, K.%
, Wang, J.%
, Lin, C\BHBI C.%
, Liu, Z.%
\BCBL {}\ \BBA {} Wang, L.%
\end{APACrefauthors}%
\unskip\
\newblock
\APACrefYearMonthDay{2023}{}{}.
\newblock
{\BBOQ}\APACrefatitle {The dawn of lmms: Preliminary explorations with gpt-4v
  (ision)} {The dawn of lmms: Preliminary explorations with gpt-4v
  (ision)}.{\BBCQ}
\newblock
\APACjournalVolNumPages{arXiv preprint arXiv:2309.17421}{9}{}{1}.
\PrintBackRefs{\CurrentBib}

\bibitem [\protect \citeauthoryear {%
Yao%
\ \protect \BOthers {.}}{%
Yao%
\ \protect \BOthers {.}}{%
{\protect \APACyear {2023}}%
}]{%
yao2023tree}
\APACinsertmetastar {%
yao2023tree}%
\begin{APACrefauthors}%
Yao, S.%
, Yu, D.%
, Zhao, J.%
, Shafran, I.%
, Griffiths, T\BPBI L.%
, Cao, Y.%
\BCBL {}\ \BBA {} Narasimhan, K.%
\end{APACrefauthors}%
\unskip\
\newblock
\APACrefYearMonthDay{2023}{}{}.
\newblock
{\BBOQ}\APACrefatitle {Tree of thoughts: Deliberate problem solving with large
  language models} {Tree of thoughts: Deliberate problem solving with large
  language models}.{\BBCQ}
\newblock
\APACjournalVolNumPages{arXiv preprint arXiv:2305.10601}{}{}{}.
\PrintBackRefs{\CurrentBib}

\bibitem [\protect \citeauthoryear {%
Zellers%
, Holtzman%
, Bisk%
, Farhadi%
\BCBL {}\ \BBA {} Choi%
}{%
Zellers%
\ \protect \BOthers {.}}{%
{\protect \APACyear {2019}}%
}]{%
zellers2019hellaswag}
\APACinsertmetastar {%
zellers2019hellaswag}%
\begin{APACrefauthors}%
Zellers, R.%
, Holtzman, A.%
, Bisk, Y.%
, Farhadi, A.%
\BCBL {}\ \BBA {} Choi, Y.%
\end{APACrefauthors}%
\unskip\
\newblock
\APACrefYearMonthDay{2019}{}{}.
\newblock
{\BBOQ}\APACrefatitle {Hellaswag: Can a machine really finish your sentence?}
  {Hellaswag: Can a machine really finish your sentence?}{\BBCQ}
\newblock
\APACjournalVolNumPages{arXiv preprint arXiv:1905.07830}{}{}{}.
\PrintBackRefs{\CurrentBib}

\bibitem [\protect \citeauthoryear {%
Zheng%
\ \protect \BOthers {.}}{%
Zheng%
\ \protect \BOthers {.}}{%
{\protect \APACyear {2023}}%
}]{%
zheng2023judging}
\APACinsertmetastar {%
zheng2023judging}%
\begin{APACrefauthors}%
Zheng, L.%
, Chiang, W\BHBI L.%
, Sheng, Y.%
, Zhuang, S.%
, Wu, Z.%
, Zhuang, Y.%
\BDBL {}others%
\end{APACrefauthors}%
\unskip\
\newblock
\APACrefYearMonthDay{2023}{}{}.
\newblock
{\BBOQ}\APACrefatitle {Judging LLM-as-a-judge with MT-Bench and Chatbot Arena}
  {Judging llm-as-a-judge with mt-bench and chatbot arena}.{\BBCQ}
\newblock
\APACjournalVolNumPages{arXiv preprint arXiv:2306.05685}{}{}{}.
\PrintBackRefs{\CurrentBib}

\end{thebibliography}

\begin{appendices}

\section*{Appendices}

\section{Example Full Prompt}

\label{app:full_prompt}

\noindent Your task is to predict the likely emotional responses of a character in this dialogue:

\noindent Cecilia: You know, your words have power, Brandon. More than you might think.\\
Brandon: I'm well aware, Cecilia. It's a critic's job to wield them.\\
Cecilia: But do you understand the weight of them? The lives they can shatter?\\
Brandon: Art is not for the faint-hearted. If you can't handle the critique, you're in the wrong industry.\\
Cecilia: It's not about handling criticism, Brandon. It's about understanding the soul of the art. You dissect it like a cold, lifeless body on an autopsy table.\\
\noindent [End dialogue]

\noindent At the end of this dialogue, Brandon would feel...\\
Offended\\
Empathetic\\
Confident\\
Dismissive

\noindent Give each of these possible emotions a score from 0-10 for the relative intensity that they are likely to be feeling each. Then critique your answer by thinking it through step by step. Finally, give your revised scores.

\noindent You must output in the following format, including headings (of course, you should give your own scores), with no additional commentary:

\noindent First pass scores:

\noindent Offended: \textless score\textgreater\\
Empathetic: \textless score\textgreater\\
Confident: \textless score\textgreater\\
Dismissive: \textless score\textgreater\\

\noindent Critique:\\
\noindent \textless your critique here\textgreater\\

\noindent Revised scores:

\noindent Offended: \textless revised score\textgreater\\
Empathetic: \textless revised score\textgreater\\
Confident: \textless revised score\textgreater\\
Dismissive: \textless revised score\textgreater\\

\noindent[End of answer]

\noindent Remember: zero is a valid score, meaning they are likely not feeling that emotion. You must score at least one emotion \textgreater 0.\\

\hrule

The full set of questions is available at: \href{https://github.com/EQ-bench/EQ-Bench}{https://github.com/EQ-bench/EQ-Bench}.

\newpage

\section{Score Correlation Matrix 1}

\label{app:score_matrix1}

\begin{table}[htbp]
  \centering
  \caption{Score Correlation Matrix: EQ-Bench vs. Arena ELO, MT-bench \& MMLU}
    \begin{tabular}{lM{2cm}M{1.5cm}M{1.5cm}M{1.5cm}M{1.5cm}}
    \toprule
    \textbf{Model} & \textbf{EQ Bench} & \textbf{Arena Elo} & \textbf{MT-bench} & \textbf{MMLU} & \textbf{Alpaca Eval} \\
    \midrule
    meta-llama/Llama-2-7b-chat-hf & 25.43 & 985   & 6.27  & 45.8  & 71.37 \\
    mistralai/Mistral-7B-Instruct-v0.1 & 28.66 & 1031  & 6.84  & --- & 54.52 \\
    meta-llama/Llama-2-13b-chat-hf & 32.34 & 1012  & 6.65  & 53.6  & 81.09 \\
    lmsys/vicuna-33b-v1.3 & 36.52 & 1092  & 7.12  & 59.2  & 88.99 \\
    gpt-3.5-0613 & 49.17 & 1115  & 7.94  & 70    & 89.39 \\
    meta-llama/Llama-2-70b-chat-hf & 51.56 & 1051  & 6.86  & 63    & 92.66 \\
    gpt-4-0613 & 62.94 & 1181  & 8.99  & 86.4  & 95.28 \\
    openchat/openchat-3.5 & 37.08 & ---   & 7.81  & 64.3  & --- \\
    text-davinci-001 & 15.19 & ---   & ---   & ---   & --- \\
    text-davinci-002 & 43.73 & ---   & ---   & ---   & --- \\
    text-davinci-003 & 39.44 & ---   & ---   & ---   & --- \\
    gpt-3.5-turbo-0301 & 49.52 & 1103   & 7.94  & 70   & --- \\
    gpt-4-0314 & 60.73 & ---   & ---   & ---   & --- \\
    Claude 1 & --- & 1146   & 7.9   & 77   & --- \\
    Claude 2 & 52.14 & 1134  & 8.06   & 78.5   & 91.36 \\
    Koala 13B & 24.92 & 955   & 5.35   & 44.7   & --- \\
    Oasst-pythia-12B & 7.69 & 884   & 4.32   & 27   & --- \\
    Dolly-V2-12B & --- & 810   & 3.28   & 25.7   & --- \\
    Alpaca 13B & --- & 893  & 4.53   & 48.1   & --- \\
    ChatGLM 6B & --- & 871   & 4.5   & 36.1   & --- \\
    lmsys/vicuna-13b-v1.1 & 32.85 & ---   & ---  & 51.9 & --- \\
    HuggingFaceH4/zephyr-7b-alpha & 38.79 & --- & --- & 61.39 & --- \\
    Qwen/Qwen-14B-Chat & 43.76 & ---  & ---   & 67.7 & --- \\
    migtissera/Synthia-70B-v1.2b & 49.99 & --- & --- & 70.13 & --- \\
    lmsys/vicuna-7b-v1.1 & 22.24 & --- & --- & 45.63 & --- \\
    \bottomrule
    \end{tabular}%
\end{table}%
\noindent Data sources: Chatbot Arena Leaderboard \citep{chatbot_arena_leaderboard}, MMLU \citep{hendrycks2020measuring}, Chatbot Arena ELO \citep{zheng2023judging}, AlpacaEval \citep{li2023alpacaeval}, MT-bench \citep{zheng2023judging}

\newpage

\section{Score Correlation Matrix 2}

\label{app:score_matrix2}

\begin{table}[htbp]
  \centering
  \caption{Score Correlation Matrix: EQ-Bench vs. Arena ELO, MT-bench \& MMLU}
    \begin{tabular}{lM{2cm}M{1.5cm}M{1.5cm}M{1.5cm}M{1.5cm}}
    \toprule
    \textbf{Model} & \textbf{EQ Bench} & \textbf{ARC} & \textbf{HellaSwag} & \textbf{Truthful-QA} & \textbf{SECEU EQ} \\
    \midrule
    meta-llama/Llama-2-7b-chat-hf & 25.43 & 52.9  & 78.55 & 45.57 & --- \\
    mistralai/Mistral-7B-Instruct-v0.1 & 28.66 & 54.52 & 75.63 & 56.28 & --- \\
    meta-llama/Llama-2-13b-chat-hf & 32.34 & 59.04 & 81.94 & 44.12 & --- \\
    lmsys/vicuna-33b-v1.3 & 36.52 & 62.12 & 83 & 56.16 & --- \\
    gpt-3.5-0613 & 49.17 & --- & --- & --- & --- \\
    meta-llama/Llama-2-70b-chat-hf & 51.56 & 64.59 & 85.88 & 52.8 & --- \\
    gpt-4-0613 & 62.94 & --- & --- & --- & --- \\
    openchat/openchat-3.5 & 37.08 & ---   & --- & 59.1 & --- \\
    text-davinci-001 & 15.19 & ---   & --- & --- & 107 \\
    text-davinci-002 & 43.73 & ---   & --- & ---  & 91 \\
    text-davinci-003 & 39.44 & ---   & --- & --- & 114 \\
    gpt-3.5-turbo-0301 & 49.52 & ---   & --- & --- & 103 \\
    gpt-4-0314 & 60.73 & ---   & --- & --- & 117 \\
    Claude 1 & --- & ---   & ---   & ---   & 106 \\
    Claude 2 & 52.14 & ---   & ---   & ---   & --- \\
    Koala 13B & 24.92 & ---   & ---   & ---   & 83 \\
    Oasst-pythia-12B & 7.69 & ---   & ---   & ---   & 107 \\
    Dolly-V2-12B & --- & ---   & ---   & ---   & 98 \\
    Alpaca 13B & --- & ---   & ---   & ---   & 104 \\
    ChatGLM 6B & --- & ---   & ---   & ---   & 94 \\
    lmsys/vicuna-13b-v1.1 & 32.85 & 52.73 & 80.14 & 52.08 & 105 \\
    HuggingFaceH4/zephyr-7b-alpha & 38.79 & 61.04 & 84.04 & 57.9 & --- \\
    Qwen/Qwen-14B-Chat & 43.76 & 58.28 & 83.99 & 49.43 & --- \\
    migtissera/Synthia-70B-v1.2b & 49.99 & 70.48 & 86.98 & 58.64 & --- \\
    lmsys/vicuna-7b-v1.1 & 22.24 & 53.67 & 77.46 & 48.94 & --- \\    
    \bottomrule
    \end{tabular}%
\end{table}%
\noindent Data sources: HellaSwag \citep{zellers2019hellaswag}, ARC \citep{clark2018think}, TruthfulQA \citep{lin2021truthfulqa} and SECEU \citep{wang2023emotional}.

\newpage

\section{SECEU EQ Scores vs. EQ-Bench scores}

\begin{table}[h]
\label{app:score_comparison}
\centering
\caption{Comparison of Original and Normalised Scores for SECEU EQ vs. EQ-Bench. Data sources: SECEU \citep{wang2023emotional}.}
\begin{tabular}{l>{\raggedright\arraybackslash}p{2.2cm}>{\raggedright\arraybackslash}p{1.8cm}>{\raggedright\arraybackslash}p{1.8cm}>{\raggedright\arraybackslash}p{1.8cm}}
\hline
\textbf{Model} & \textbf{SECEU EQ} & \textbf{EQ-Bench} & \textbf{SECEU EQ Normalised} & \textbf{EQ-Bench Normalised} \\
\hline
Oasst & 107 & 7.69 & 70.59 & 26.30 \\
text-curie-001 & 102 & 13.55 & 55.88 & 34.44 \\
text-davinci-001 & 107 & 15.19 & 70.59 & 36.72 \\
Koala 13b & 83 & 24.92 & 0.00 & 50.24 \\
Vicuna 13b & 105 & 32.85 & 64.71 & 61.26 \\
text-davinci-003 & 114 & 39.44 & 91.18 & 70.42 \\
text-davinci-002 & 91 & 43.73 & 23.53 & 76.38 \\
gpt-3.5-turbo-0301 & 103 & 49.52 & 58.82 & 84.42 \\
gpt-4-0314 & 117 & 60.73 & 100.00 & 100.00 \\
\hline
\end{tabular}

\end{table}

\newpage

\section{EQ-Bench Repeatability \& First Pass vs. Revised Scores}

\begin{table}[h]
\centering
\caption{EQ-Bench First Pass vs. Revised Scores.}
\label{app:repeatability}
\begin{tabular}{lcc}
\hline
Model & First Pass Score & Revised Score \\
\hline
gpt-3.5-turbo-0301 & 47.18 & 48.72 \\
gpt-3.5-turbo-0301 & 46.44 & 49.56 \\
gpt-3.5-turbo-0301 & 47.95 & 48.46 \\
gpt-3.5-turbo-0613 & 47.60 & 48.99 \\
gpt-3.5-turbo-0613 & 48.98 & 47.61 \\
gpt-3.5-turbo-0613 & 48.94 & 49.75 \\
gpt-4-1106-preview & 53.45 & 56.94 \\
gpt-4-1106-preview & 53.31 & 56.83 \\
gpt-4-1106-preview & 54.34 & 56.40 \\
llama2-7b-chat & 24.61 & 25.54 \\
llama2-7b-chat & 24.68 & 25.47 \\
llama2-7b-chat & 24.61 & 25.68 \\
mistral-openorca-7b & 45.25 & 42.38 \\
mistral-openorca-7b & 45.02 & 42.90 \\
mistral-openorca-7b & 45.00 & 43.66 \\
qwen-14b & 43.31 & 36.94 \\
qwen-14b & 44.55 & 41.28 \\
qwen-14b & 44.89 & 42.18 \\
synthia-70b & 54.99 & 53.77 \\
synthia-70b & 55.12 & 55.37 \\
synthia-70b & 55.30 & 55.06 \\
text-davinci-001 & 11.30 & 18.25 \\
text-davinci-001 & 8.26 & 16.56 \\
text-davinci-001 & 8.83 & 15.42 \\
text-davinci-002 & 32.19 & 37.20 \\
text-davinci-002 & 32.78 & 39.91 \\
text-davinci-002 & 32.26 & 40.07 \\
text-davinci-003 & 44.64 & 43.14 \\
text-davinci-003 & 44.31 & 43.01 \\
text-davinci-003 & 44.71 & 43.72 \\
vicuna-7b-1.1 & 21.75 & 21.64 \\
vicuna-7b-1.1 & 21.76 & 21.21 \\
vicuna-7b-1.1 & 22.03 & 22.07 \\
\hline
\end{tabular}
\end{table}

\end{appendices}

\end{document}